\documentclass{article}
\usepackage{ijcai25}

\usepackage{times}
\usepackage{soul}
\usepackage{url}
\usepackage[hidelinks]{hyperref}
\usepackage[utf8]{inputenc}
\usepackage[small]{caption}
\usepackage{graphicx}
\usepackage{amsmath}
\usepackage{amsthm}
\usepackage{booktabs}
\usepackage{algorithm}
\usepackage{algorithmic}
\usepackage[switch]{lineno}

\title{FHBench: Towards Efficient and Personalized Federated Learning for Multimodal Healthcare}

\author{
Penghao Wang$^1$
\and
Qian Chen$^2$ \and
Teng Zhang$^2$ \and
Yingwei Zhang$^2$ \and
Wang Lu$^2$ \and
Yiqiang Chen$^2$ \\
\affiliations
$^1$ Independent Researcher \\
$^2$ Institute of Computing Technology, Chinese Academy of Sciences 
}

\begin{document}

\maketitle

\begin{abstract}
Federated Learning (FL) has emerged as an effective solution for multi-institutional collaborations without sharing patient data, offering a range of methods tailored for diverse applications. However, real-world medical datasets are often multimodal, and computational resources are limited, posing significant challenges for existing FL approaches. Recognizing these limitations, we developed the Federated Healthcare Benchmark(FHBench), a benchmark specifically designed from datasets derived from real-world healthcare applications. FHBench encompasses critical diagnostic tasks across domains such as the nervous, cardiovascular, and respiratory systems and general pathology, providing comprehensive support for multimodal healthcare evaluations and filling a significant gap in existing benchmarks. Building on FHBench, we introduced Efficient Personalized Federated Learning with Adaptive LoRA (EPFL), a personalized FL framework that demonstrates superior efficiency and effectiveness across various healthcare modalities. Our results highlight the robustness of FHBench as a benchmarking tool and the potential of EPFL as an innovative approach to advancing healthcare-focused FL, addressing key limitations of existing methods. Code is available at \url{https://github.com/wph6/FHBench}

\end{abstract}

\section{Introduction}

Federated Learning (FL)\cite{fedavg} enables collaborative model training across decentralized clients while ensuring data privacy by keeping raw data local. This privacy-preserving approach is particularly valuable in healthcare, where sensitive data and strict confidentiality are critical\cite{xu2021federated}. Nevertheless, healthcare data is inherently heterogeneous, distributed, and multimodal, posing significant challenges to conventional FL methods. For example, patient demographics, disease prevalence, and imaging protocols often vary across institutions, leading to highly non-IID data distributions. A specific example is time-series data such as electrocardiograms (ECG), where variations in patient populations and recording conditions exacerbate data heterogeneity, limiting the performance of naive federated models\cite{minchole2019ecg}. Addressing these issues requires tailored evaluation methods that capture the complexities of healthcare data. Existing benchmarks such as LEAF\cite{caldas2018leaf}, often fall short in reflecting real-world medical tasks, which involve multimodal data such as imaging and audio. Additionally, the diverse and non-IID nature of healthcare datasets further complicates benchmark design. Without comprehensive benchmarks, it is difficult to evaluate the effectiveness of federated and personalized learning methods in addressing these unique challenges.This concern is further echoed in recent discussions on the limitations of federated foundation models, which highlight the need for dedicated evaluation frameworks in complex domains like healthcare \cite{fan2025ten}.

Beyond evaluation challenges, the complexity of healthcare tasks also demands more advanced models that can extract meaningful representations from diverse and multimodal data. Large pre-trained models, such as GPT\cite{radford2018gpt} and BERT\cite{devlin2018bert}, have demonstrated exceptional capabilities in general-purpose feature extraction, making them promising candidates for improving FL performance in healthcare. These pre-trained models offer the potential to bridge gaps in representation learning for multimodal and heterogeneous data, as demonstrated in biomedical FL scenarios like Buffalo \cite{yan2024buffalo}, which combines vision-language modeling with federated learning. However, their high computational and fine-tuning costs often conflict with the decentralized and resource-limited nature of FL.

To tackle these challenges, Parameter-Efficient Fine-Tuning (PEFT) methods, such as prefix tuning\cite{li2021prefix} and LoRA (Low-Rank Adaptation)\cite{hu2022lora}, have been introduced to address this challenge. LoRA, in particular, improves fine-tuning efficiency through low-rank updates, enabling cost-effective model adaptation. Its lightweight design makes it well-suited for FL, supporting efficient and privacy-preserving decentralized model updates. This integration opens new opportunities for addressing the challenges of non-IID data and resource constraints in healthcare FL. Building on these observations, Our study aims to answer the following key questions:
\begin{figure*}[t] 
    \centering 
    \includegraphics[width=\textwidth]{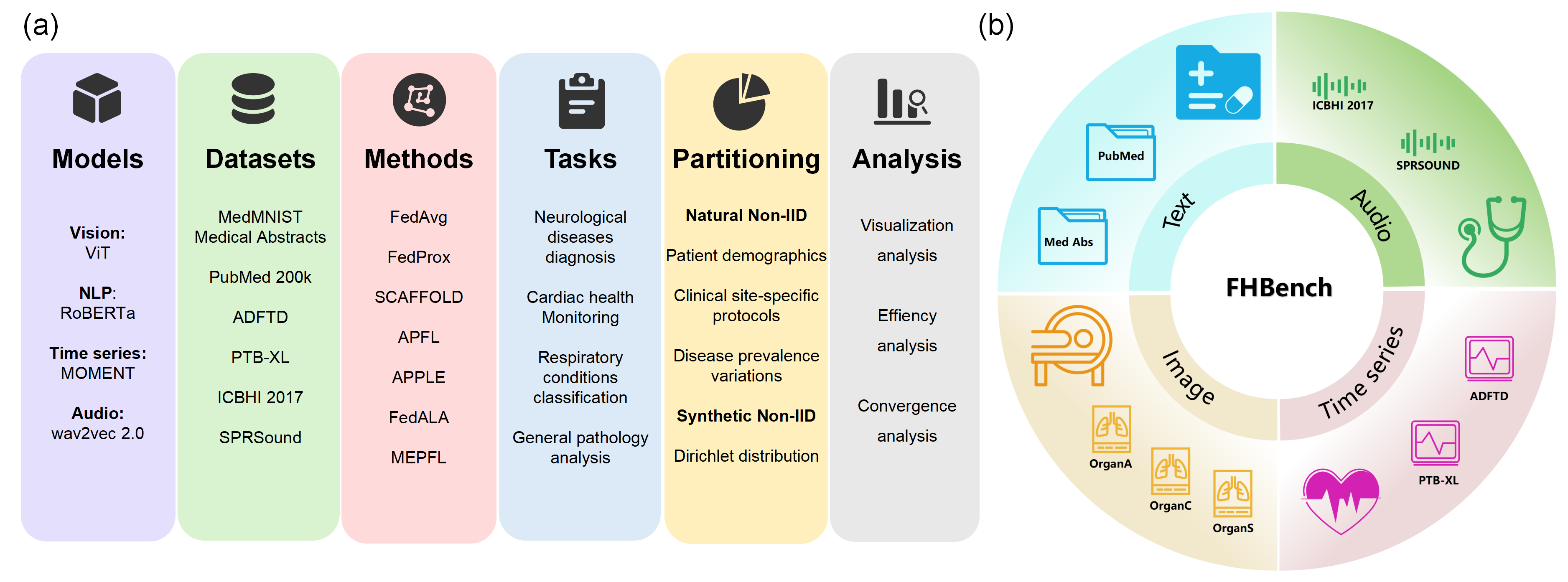}
    \caption{(a) Components of the FHBench framework.(b) Dataset composition in FHBench across multiple modalities.} 
    \label{pie} 
\end{figure*} 

\textbf{Challenges: How Do Existing Benchmarks and Methods Fall Short?}
In Section \ref{related}, we analyze the limitations of existing benchmarks and methods in healthcare FL. Current benchmarks fail to capture the complexity of multimodal medical data and realistic non-IID distributions. Additionally, while large pre-trained models hold promise for improving medical FL, their computational and communication inefficiencies in decentralized settings remain unresolved. These gaps highlight the need for tailored benchmarks and efficient methods to address the unique challenges of healthcare in FL.

\textbf{Insights: What Makes Benchmarking in Healthcare Unique?}
In Section \ref{benchmark}, we introduce the FHBench, designed to address the challenges posed by healthcare data. Covering diverse diagnostic tasks across the nervous, cardiovascular, and respiratory systems, as well as general pathology, FHBench incorporates both natural and synthetic non-IID partitions. This design captures the inherent heterogeneity of real-world medical scenarios, providing a robust and practical platform for evaluating FL methods.

\textbf{Advancements: How Can FL Methods be Designed for Efficiency in Healthcare?} In Section \ref{method}, we introduce \textbf{E}fficient \textbf{P}ersonalized \textbf{F}ederated \textbf{L}earning with Adaptive LoRA (EPFL), a personalized FL framework that integrates Low-Rank Adaptation (LoRA) to optimize computational and communication efficiency. By employing similarity-weighted aggregation, EPFL balances global knowledge sharing with client-specific adaptation. Its lightweight design enables the integration of large pre-trained models, enhancing representation learning for multimodal medical tasks. Evaluations on FHBench validate its effectiveness in addressing healthcare-specific challenges.

\section{Related Work}\label{related}

\subsection{FL in Healthcare}

Healthcare is a privacy-sensitive domain that benefits significantly from FL, which enables collaborative model training without sharing raw data. 
FL has been applied to a wide range of healthcare data and applications, including brain tumor segmentation\cite{li2019brain} and drug discovery\cite{oldenhof2023drug}. However, the heterogeneity, non-IID distributions, and multimodal nature of healthcare data pose significant challenges to the efficiency and effectiveness of traditional FL methods. To address these issues, Personalized federated learning (PFL) has emerged as a promising approach, tailoring models to individual client data distributions. By focusing on personalization, PFL seeks to enhance performance across diverse datasets while maintaining strict data privacy.

Several PFL algorithms have been developed to address data heterogeneity in healthcare. For instance, FedAP\cite{lu2022fedap} personalizes batch normalization layers to adapt to non-IID data, improving performance in healthcare settings while ensuring privacy. Similarly, MetaFed\cite{chen2023metafed} utilizes cyclic knowledge distillation to create personalized models for each client, achieving enhanced privacy and accuracy in diverse healthcare applications.

\subsection{Efficicent medical FL with Pre-trained Models}
FL has been increasingly applied to training large pre-trained models due to its privacy-preserving nature, enabling collaborative model development without sharing raw data. While methods like OpenFedLLM\cite{ye2024openfedllm} effectively apply FL to large language models (LLMs), challenges such as high communication costs, computational overhead, and scalability persist. Recent advancements in Parameter-Efficient Fine-Tuning (PEFT) offer promising solutions. For example, FedPepTAO\cite{che2023FedPepTAO} uses prompt tuning and adaptive optimization to efficiently fine-tune LLMs, reducing resource usage while maintaining performance. Similarly, FedPETuning\cite{zhang2023fedpetuning} benchmarks PEFT methods in FL, demonstrating improvements in privacy, communication efficiency, and model performance.

Among the PEFT methods, Low-Rank Adaptation (LoRA) has gained significant attention for its efficiency and simplicity. This efficiency makes LoRA particularly well-suited for FL, where resource constraints and communication efficiency are critical considerations. FLoRA\cite{nguyen2024flora} utilizes FL and LoRA to train vision-language models (VLMs) in a privacy-preserving and computationally efficient manner. Similarly, FDLoRA\cite{qi2024fdlora} personalizes large language models (LLMs) in FL using a dual LoRA tuning approach. While effective, its dual-module design adds complexity, potentially limiting adaptability in resource-constrained healthcare scenarios.

\subsection{Medicial FL benchmark}

Benchmarks are essential for advancing FL research, offering standardized datasets and evaluation protocols to rigorously assess algorithm effectiveness. In the healthcare domain, the development of FL benchmarks is particularly important due to the unique challenges of medical data, such as heterogeneity, non-IID distributions, and strict privacy requirements.

Several benchmarks have been proposed to evaluate FL methods in general settings. For instance, LEAF offers datasets for natural language processing and image classification tasks. FedNLP\cite{lin2022fednlp} serves as a benchmark for assessing FL methods on tasks such as text classification, sequence tagging, and question answering, providing valuable insights into the performance of FL models on language-specific tasks. FedMultimodal\cite{feng2023fedmultimodal} extends the scope of FL benchmarking to multimodal FL, encompassing a variety of applications and data modalities.

While these benchmarks have advanced FL research in general domains, they fall short of addressing the unique challenges posed by medical data. Medical tasks often encompass multiple modalities, such as imaging, text, time-series signals, and audio data, reflecting the complexity of real-world diagnostics. These characteristics highlight the need for benchmarks specifically designed to evaluate FL methods in healthcare contexts. Moreover, in resource-constrained medical real-world scenarios, there is a growing demand for methods that not only harness the prior knowledge of large pre-trained models but also ensure computational and communication efficiency.

\begin{figure*}[t] 
    \centering 
    \includegraphics[width=\textwidth]{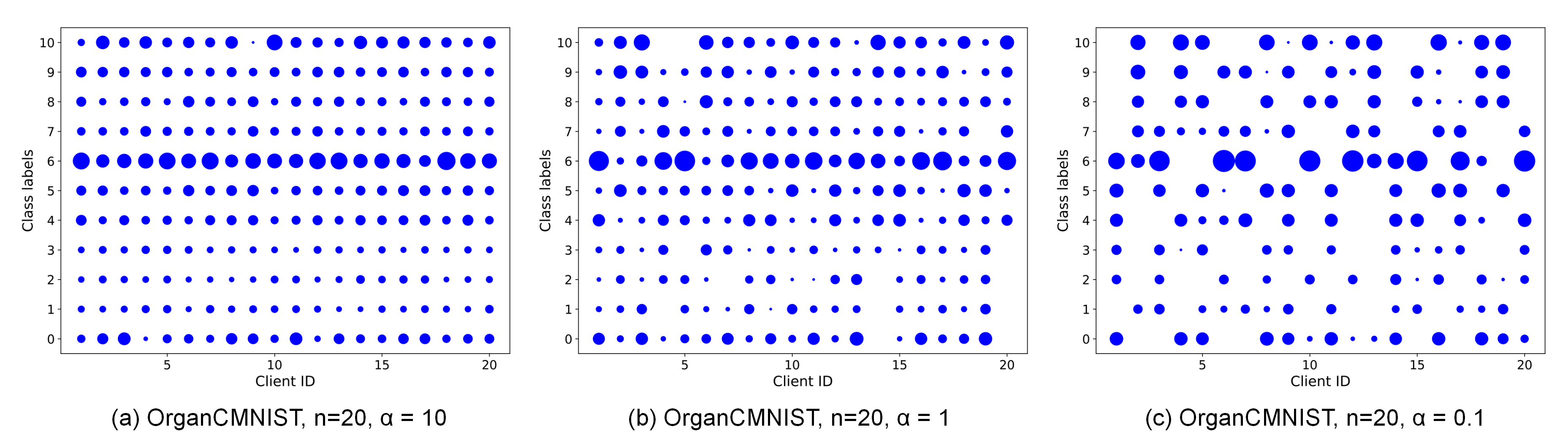}
    \caption{Visualization of Non-IID Data Partitions Across Clients at Different Levels of Heterogeneity, showing the Number of Samples per Class Allocated to Each Client (Indicated by Dot Sizes)} 
    \label{iid} 
\end{figure*} 

\section{FHBench}\label{benchmark}
Designing a dedicated benchmark for medical FL is essential for advancing research in this domain. An effective benchmark not only facilitates the comprehensive evaluation of FL methods but also captures the unique challenges in healthcare, including non-IID data distributions, multimodal tasks, and strict privacy requirements. In this section, we introduce the FHBench, a practical and versatile platform specifically designed to evaluate the performance of FL methods in realistic medical scenarios.

\subsection{Objectives}
In PFL, we assume the presence of \( N \) distinct clients, denoted as \( \{C_1, C_2, \dots, C_N\} \). Each client has its own dataset \( \{D_1, D_2, \dots, D_N \}\), where the datasets are non-IID. We assume that the dataset $D_i$ is not large enough for clients to accurately approximate a solution to their local problem on their own. Each client maintains its own model parameters, denoted as \( \{\theta_i\}_{i=1}^N \), all of which are fine-tuned from the same pre-trained model. The objective is to aggregate information from all clients to learn suitable model parameters \( f_i \) for each client, based on its local dataset \( D_i \), while ensuring no private data leakage. This is achieved by minimizing the global objective function:

\begin{equation}
\min F(\mathbf{\Theta}) := \min \frac{1}{n} \sum_{i=1}^{n} f_i(\mathbf{\theta}_i) 
\end{equation}

where \( \mathbf{\Theta} \) represents the collection of model parameters from all clients, and \( \mathbf{\theta}_i \) represents the model parameters of the \( i \)-th client. The global objective is to minimize the overall performance loss by optimizing the local objectives \( f_i \) for each client.

Each client's local objective \( f_i \) is defined as the average loss over its dataset \( D_i \), calculated as follows:

\begin{equation}
f_i(\mathbf{\theta_i}) := \frac{1}{|D_i|} \sum_{\xi \in D_i} \ell(\mathbf{\theta}_i, \xi) 
\end{equation}

\( |D_i| \) represents the size of the dataset for client \( i \), \( \ell \) is the loss function, typically a cross-entropy loss and \( \xi \) denotes a data point in \( D_i \). This approach ensures that each client’s model training objective is to minimize the average loss across its local data.

\subsection{Components}
\textbf{Datasets.} We have curated a suite of realistic federated datasets for FHBench, focusing on datasets that (1) are derived from authentic medical applications, reflecting real-world diagnostic tasks; (2) represent diverse medical modalities, including imaging, text, time-series, and audio; and (3) encompass critical diagnostic challenges across major medical domains, including the nervous system, cardiovascular system, respiratory system, and general pathology. Currently, FHBench consists of the following datasets:

\begin{itemize}
    \item \textbf{MedMnist}\cite{yang2021medmnist1,yang2023medmnist2}: A standardized collection of biomedical image datasets designed for multi-organ diagnostics, including high-resolution 2D and 3D CT slices.
    \item \textbf{Medical Abstracts Text}\cite{medabs}: A corpus of medical text abstracts categorized into diverse pathological conditions, supporting disease diagnosis and knowledge extraction.
    \item \textbf{PubMed 200k RCT}\cite{dernoncourt2017pubmed}: A dataset focused on sequential sentence classification from medical abstracts, covering key sections like objectives, methods, and conclusions.
    \item \textbf{ADFTD}\cite{ADFTD1,ADFTD2}: An EEG dataset capturing brain activity from Alzheimer’s and Frontotemporal Dementia patients, aimed at neurodegenerative disorder classification.
    \item \textbf{PTB-XL}\cite{ptbxl}: An dataset representing various cardiac conditions, enabling tasks like arrhythmia detection and cardiac health monitoring.
    \item \textbf{ICBHI 2017}\cite{ICHBI}: A respiratory sound dataset designed for the classification of breathing patterns and respiratory abnormalities.
    \item \textbf{SPRSound}\cite{sprsound}: A pediatric respiratory sound dataset supporting the detection of conditions such as asthma, stridor, and chronic lung diseases in children.
\end{itemize}
For brevity, detailed descriptions of each dataset and preprocessing steps are provided in \textbf{Appendix \ref{datasets}}.
The statistical information of these datasets is presented in Table \ref{Statistical}.

\textbf{Models and Tasks.} To comprehensively evaluate the performance of EPFL on FHBench across diverse data modalities, we selected state-of-the-art backbone models specifically tailored to each data type. The following outlines the models chosen for fine-tuning and the rationale behind their selection:

\begin{itemize}
    \item \textbf{Visual Modality:}
    For image classification tasks, we use the Vision Transformer (ViT)\cite{ViT}, which processes image patches as sequences, leveraging the self-attention mechanism for superior performance on high-resolution medical images.
    \item \textbf{Text Modality:}
    For clinical text analysis, we employ RoBERTa\cite{liu2019roberta}, an optimized version of BERT\cite{devlin2018bert}, fine-tuned on larger corpora with dynamic masking to enhance its performance in text classification.
    \item \textbf{Time Series Modality:}
    For time-series data, we utilize the MOMENT\cite{goswami2024moment} model, designed to capture temporal dependencies and complex patterns, ensuring robustness in physiological signal analysis.
    \item \textbf{Audio Modality:}
    For audio classification tasks, we apply wav2vec 2.0\cite{baevski2020wav2vec}, which uses self-supervised learning to produce rich audio representations, significantly improving downstream respiratory sound analysis.
\end{itemize}

\begin{figure}[h] 
    \centering 
    \includegraphics[width=0.5\textwidth]{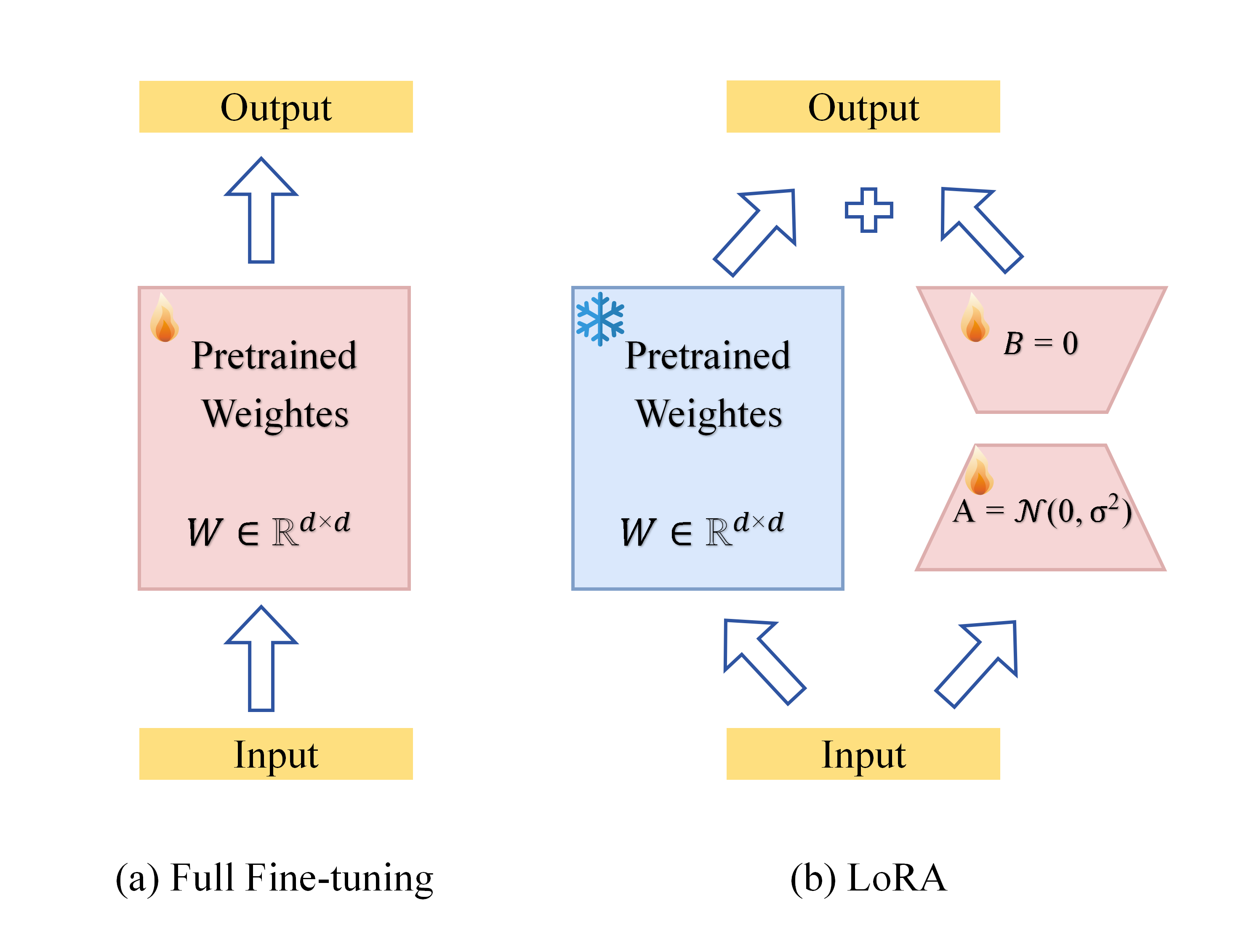}
    \caption{Comparison of Full Fine-tuning (left) and LoRA Fine-tuning (right)} 
    \label{lora} 
\end{figure} 

\begin{table*}[t]
    \centering
    \begin{tabular}{llllll}
        \hline
            Dataset  & Modalities &Disease Type  &Partition &\#Class &\#Sample   \\
        \hline
        OrganAMNIST             & Image       &General Pathologies         &Synthetic    & 11    & 58850 \\
        OrganCMNIST             & Image       &General Pathologies         &Synthetic    & 11    & 23660 \\
        OrganSMNIST             & Image       &General Pathologies         &Synthetic    & 11    & 25221 \\
        Medical Abstracts Text  & Text        &General Pathologies         &Synthetic    & 5     & 14438 \\
        PubMed 200k RCT          & Text        &General Pathologies         &Synthetic    & 5     & 195,654 \\
        ADFTD                   & time series &Neurodegenerative Diseases   &Synthetic/Natural& 3     & 69,752 \\
        PTB-XL                   & time series &Cardiac Diseases             &Synthetic/Natural& 5     & 191,400 \\
        ICBHI 2017              & Audio       &Respiratory Disorders        &Synthetic/Natural& 4     & 6898 \\
        SPRSound                & Audio       &Respiratory Disorders        &Synthetic/Natural& 7     & 9,089 \\
        \hline
    \end{tabular}
    \caption{Statistical information of datasets included in FHBench.}
    \label{Statistical}
\end{table*}

\textbf{Baselines.} We implemented an extensible experimental framework to evaluate FHBench, incorporating both classic FL methods (FedAvg, FedProx\cite{fedprox}, SCAFFOLD\cite{karimireddy2020scaffold}) and personalized FL methods (APFL\cite{deng2020APFL}, fedALA\cite{zhang2023fedala}, APPLE\cite{luo2022apple}). The framework is designed for scalability, allowing the integration of additional algorithms, datasets, and modalities. For detailed descriptions of the methods and implementation, please refer to \textbf{Appendix \ref{appendix method}}.
It offers diverse evaluation protocols and modular customization, providing a robust platform for benchmarking FL methods in healthcare. However, given the computational and communication constraints inherent in many healthcare scenarios, efficiency becomes a critical factor for practical adoption. To address these challenges, our framework integrates PEFT methods, with a focus on LoRA, a method specifically designed to reduce the resource overhead associated with adapting large pre-trained models in federated settings. LoRA achieves the weight update as $\Delta W = B A$, where $B \in \mathbf{R}^{d \times r}$ and $A \in \mathbf{R}^{r \times k}$ are low-rank matrices with $r \ll \min(d, k)$. The output for an input $x$ is computed as:

\begin{equation}
h = W_0 x + \Delta W x = W_0x + BAx 
\end{equation}
where \( W_0 \) represents the frozen pre-trained weights, and \( B A x \) captures the adaptation. As shown in Figure \ref{lora}, unlike Full Fine-Tuning, which updates all parameters and demands high memory and computation, LoRA optimizes only the low-rank B and A matrices, significantly enhancing efficiency while maintaining adaptability.

\textbf{Partitioning.} Non-IID data partitioning is essential for simulating realistic FL experiments, especially in healthcare. Medical data naturally exhibits non-IID characteristics due to factors such as patient demographics, site-specific protocols, and variations in disease prevalence. To reflect this heterogeneity in FHBench, we implement both natural and synthetic non-IID partitioning schemes tailored to different datasets, closely aligning with real-world healthcare scenarios.

For datasets like Physiological Signals (ADFTD and PTBXL) and Respiratory Sounds (ICBHI 2017 and SPRSound), the data is naturally organized by patient or recording session IDs. This structure captures variations in demographics (e.g., age, gender) and recording conditions (e.g., device settings, environmental noise), resulting in authentic non-IID distributions that mimic personalized healthcare settings.

In contrast, for tasks like General Pathology (e.g., MedMNIST and Medical Abstracts Text), where data lacks inherent client-specific partitions, we use synthetic non-IID partitioning via the Dirichlet distribution. The degree of heterogeneity is controlled by the parameter $\alpha$: lower values (e.g., $\alpha = 0.1$) create highly skewed distributions, simulating extreme heterogeneity, while higher values (e.g., $\alpha = 10$) yield more balanced splits. Figure \ref{iid} illustrates these partitions, providing a rigorous evaluation framework for FL methods by enhancing data diversity and replicating real-world challenges.

\textbf{Analysis.} While our framework ensures scalability and adaptability, a key focus is on addressing gaps in existing methods. Experimental results reveal limitations in the personalization degree, convergence speed, and parameter efficiency of current FL approaches, as discussed in Section \ref{analysis}.

Building on prior insights, Zhu et al.\shortcite{zhu2024asymmetry} identified distinct roles for the LoRA matrices: the A-matrix extracts input features, while the B-matrix generates task-specific outputs, with fine-tuning B being more effective. However, in healthcare scenarios, the inherent non-IID data distributions, limited data diversity, and strict privacy constraints complicate learning robust global feature extractors. Leveraging these characteristics, we designed EPFL to address these challenges, as detailed in the following section.

\section{EPFL}\label{method}
Due to the suboptimal performance of existing methods on standard benchmarks, we propose EPFL, a novel approach designed to enhance and improve performance. Traditional FL processes typically involve client devices training models locally, followed by the aggregation of these models on a global server to form a unified global model. However, these methods often face challenges such as inefficiency, limited scalability, and insufficient adaptability. To address these issues, EPFL introduces an adaptive LoRA matrix, enabling more effective collaboration between clients and the global server, thus achieving superior performance in FL tasks without compromising privacy and security. The aggregation formula used in each training round is given by:

\begin{equation}
x_{global}=\sum_{i=1}^{N}p_{i}x_{i}
\end{equation}
where \( x_i \) represents the model parameters from client \( i \), and \( p_i \) denotes the aggregation weight for each client's model, calculated based on a specific criterion.

Contrary to this traditional approach, EPFL exploits the unique properties of LoRA matrices for personalized aggregation. Rather than combining into a single global model, the method specifically aggregates the \( A \) matrices from the LoRA layers across clients. The full set of parameters for each client model is denoted by \(\theta_i = \phi_{A_i} \cup \phi_{B_i}\), where \(\phi_{A_i}\) and \(\phi_{B_i}\) represent the parameters of the A and B matrix layers, respectively, tailored to each client.

The \( B \) matrix, acting as a personalized training parameter, plays a pivotal role in this setup. To reduce computational overhead, we introduce a hyperparameter \( \psi \) that selectively controls the participation of LoRA layers in the distance computation. The weights for aggregating these parameters are based on the distance between the \( B \) matrices of different clients, where only selected layers are considered:
\begin{equation}
d_{ij}^{(\ell)} = \psi^{(\ell)} \cdot \| \phi_{Bi}^{(\ell)} - \phi_{Bj}^{(\ell)} \|_2,
\end{equation}
\begin{equation}
\bar{d}_{ij} = \frac{1}{\sum_{\ell=1}^L \psi^{(\ell)}} \sum_{\ell=1}^L d_{ij}^{(\ell)} 
\end{equation}
where \( L \) represents the number of layers and \( \psi^{(\ell)} \) is a binary indicator (0 or 1) specifying whether the \( \ell \)-th layer is included in the computation. Weights \( s_{ij} \) are set as the inverse of \( d_{ij} \), and normalized as follows:

\begin{equation}
\hat{s}_{ij} = \frac{1/d_{ij}}{\sum_{k=1, k \neq i}^{N} \frac{1}{d_{ik}}}, \quad \text{for } i \neq j \text{ and } d_{ij} \neq 0.
\end{equation}

To ensure stability during training, we incorporate a moving average update strategy for \(\psi^{t+1}\), with:
\begin{equation}
s_{ij} = \begin{cases}
\lambda, & \text{if } i = j, \\
(1 - \lambda) \times \hat{s}_{ij}, & \text{if } i \neq j.
\end{cases}
\end{equation}

To reduce computation overhead, we introduce a hyperparameter \( \psi \) that controls the participation of LoRA layers in the aggregation process.

Let \(\theta_i^t = \phi_{Ai}^t \cup \phi_{Bi}^t\) represent the parameters of the model from client \(i\) at round \(t\). After updating \(\theta_i^t\) using the local data from client \(i\), the clients will independently perform updates on their own data across \(t\) rounds :

\begin{equation}
    \begin{cases}
    \phi_{Ai}^{t+1}=\sum_{j=1}^{N}s_{ij}\phi_{Aj}^{t} \\
    \phi_{Bi}^{t+1}=\phi_{Bi}^{t} & 
    \end{cases}
\end{equation}

\begin{algorithm}[tb]
    \caption{EPFL}
    \label{alg:algorithm}
    \textbf{Input}:
    $N$: number of clients,
    $\{D_i\}_{i=1}^N$:datasets of $N$ clients,
    $\theta_0$:initial model parameter,  
    $T$: number of training rounds,
    $\phi$: hyperparameter controlling layer participation,
    $\delta$: local learning rate\\
    \textbf{Output}: 
    client models' parameters $\{\theta_i\}_{i=1}^N$
    
    \begin{algorithmic}[1] 
    \FOR{Round $t = 1, \ldots, T$}
        \FOR{Client $i = 1, \ldots, N$}
            \STATE $\triangleright$ \textbf{Local training}
            \STATE $\theta_i^{t+1} \leftarrow \theta_i^t - \delta \nabla [\mathcal{L}(\Theta_i^t; D_i)]$
        \ENDFOR
        
        \STATE $\triangleright$ \textbf{Compute distance matrices}
        
        \STATE $d_{ij}^{(\ell)} = \psi^{(\ell)} \cdot \| \phi_{Bi}^{(\ell)} - \phi_{Bj}^{(\ell)} \|_2,$
        
        \STATE $\bar{d}_{ij} = \frac{1}{\sum_{\ell=1}^L \psi^{(\ell)}} \sum_{\ell=1}^L d_{ij}^{(\ell)},$
        
        \STATE $\triangleright$ \textbf{Compute aggregation weights \( w_{ij} \) }
        
        \STATE $s_{ij} = \begin{cases}
                \lambda, & \text{if } i = j, \\
                (1 - \lambda) \times \hat{s}_{ij}, & \text{if } i \neq j.
                \end{cases}$

        \STATE $\triangleright$ \textbf{Updating $\theta_i^t$ for each client}
        
        \STATE $\begin{cases}
                \phi_{Ai}^{t+1}=\sum_{j=1}^{N}s_{ij}\phi_{Ai}^{t} \\
                \phi_{Bi}^{t+1}=\phi_{Bi}^{t} & 
                \end{cases}$
                
    \ENDFOR
    \RETURN$\theta_1, \theta_2, \ldots, \theta_N$
        
    \end{algorithmic}
\end{algorithm}

\section{Experiments}

\begin{table*}[t]
    \centering
    \begin{tabular}{c|c c c c c c c}
    \toprule
    \textbf{Methods} & \multicolumn{1}{c}{\textbf{FedAvg}} & \textbf{FedProx} & \textbf{SCAFFOLD} & \textbf{APFL} & \textbf{APPLE} & \textbf{FedALA} & \textbf{EPFL(Ours)} \\
    \midrule
    OrganSMNIST & 67.024  & 66.598          & 66.262         & 67.525 &  68.554 & 68.419  & \textbf{71.445}\\
    OrganAMNIST & 86.818 & 86.868          & 85.44           & 83.591 & 86.539  &86.974   & \textbf{88.016}\\
    OrganCMNIST & 76.070 & 75.983          & 73.611          & 73.074 & 75.969  & 74.834  &\textbf{79.320}\\ 
    \midrule
    Medical Abstracts Text    & 73.298  & 72.785         & 67.257          & 54.340 & 71.592 & 73.835   &\textbf{74.294}\\
    PubMed200k RCT     & 85.53  & 85.35        & 80.947          & 56.330 & 85.651 &86.032   &\textbf{86.362}\\ 
    \midrule
    ADFTD       & 63.391 & 65.578          & 58.219          & 66.420 & 58.791 & 63.423   &\textbf{67.005}\\
    PTB-XL       & 74.489 & 75.679          & 74.321          & 73.618  & 75.371 & 75.712   &\textbf{75.741}\\ 
    \midrule
    ICBHI       & 67.071 & 59.979          & 74.061          & 75.313 & 73.806 & 67.650    &\textbf{75.713}\\
    SPRS        & 77.786 & 77.041          & 77.443         & 75.908  & 77.956 & 76.645   & \textbf{78.191}                   \\ 
    \bottomrule
    \end{tabular}
    \caption{Performance Comparison on FHBench Datasets}
    \label{main}
\end{table*}

We evaluate our proposed method on FHBench by comparing its performance against six established FL algorithms. To simulate a scenario of data insufficiency typical in federated settings, we select appropriate portions of datasets using the same random seed for all experiments. Further, we utilize a Dirichlet distribution to create disjoint non-IID client datasets. We partition the data into training, testing, and validation sets with a ratio of 4:3:3. The Dirichlet distribution parameter $\alpha$ is set to 0.1 to reflect the default heterogeneous setting. Our experiments are conducted with 20 clients over 200 training rounds using cross-entropy loss and the SGD optimizer. The similarity across all layers is computed by default. All instances of the hyperparameter $\psi^{(\ell)}$ are set to 1 by default to calculate the similarity across all layers. Additionally, for LoRA, the rank is fixed at 8 across all experiments to ensure consistent parameter efficiency. We run three trials to record the average results.

\subsection{Results}
The classification results of each algorithm on our benchmark are presented in Table \ref{main}. 

\textbf{Overall Performance:} Our proposed method consistently outperforms other algorithms across multiple data modalities, highlighting its robustness and effectiveness. Notably, on visual datasets, it achieves significant performance gains, further validating the efficacy of the adaptive LoRA matrix and its ability to enhance model convergence in federated learning.

\textbf{Challenges with Specific Modalities:} On certain modalities, such as the ADFTD time-series dataset, the performance improvements of our method are less pronounced. The results across all FL algorithms remain relatively low and comparable. This is likely due to the dataset's limited class diversity, pronounced non-IID characteristics, and small data volume. These factors contribute to overfitting on client-specific data and exacerbate non-IID effects, such as divergent learning paths and gradient conflicts during model aggregation, ultimately impeding the global model's convergence.

\subsection{Ablation study}
\begin{figure}[htb] 
    \raggedright
    \includegraphics[width=0.5\textwidth]{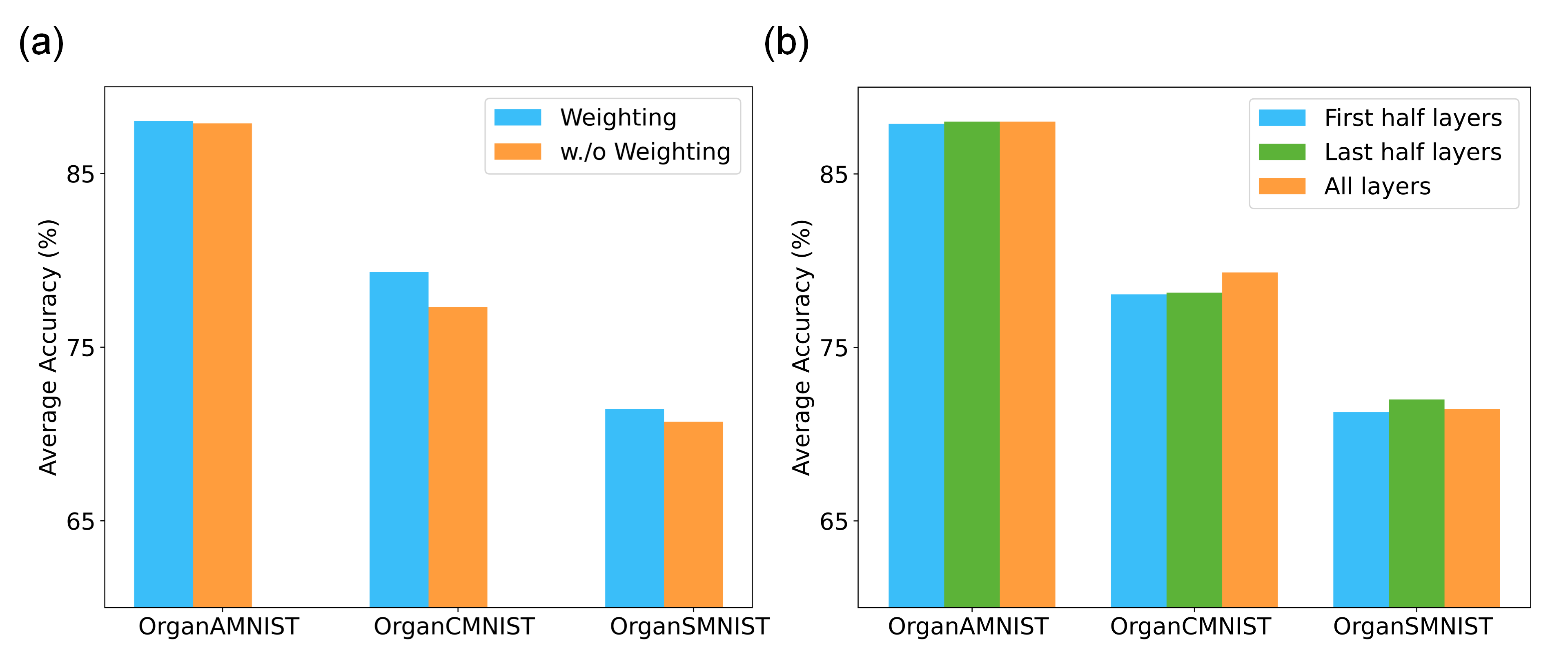}
    \caption{Impact of Similarity-Weighted Aggregation and Parameter Selection} 
    \label{abaltion} 
\end{figure}

To demonstrate the effect of a weighting scheme that accounts for the similarities among different clients, we compared the average accuracies on the OrganAMNIST, OrganCMNIST, and OrganSMNIST datasets between experiments using personalized weights and simple averaging. Each experiment was repeated three times to ensure consistency, and the results are presented in Figure \ref{abaltion}(a).

The results demonstrate that the similarity-weighted aggregation scheme consistently outperformed simple averaging across all datasets. Notably, the improvements were more pronounced on datasets with smaller sample sizes, such as OrganCMNIST and OrganSMNIST, highlighting the effectiveness of leveraging client similarities to improve personalization in data-scarce scenarios.

\begin{figure}[htb] 
    \centering 
    \includegraphics[width=0.35\textwidth]{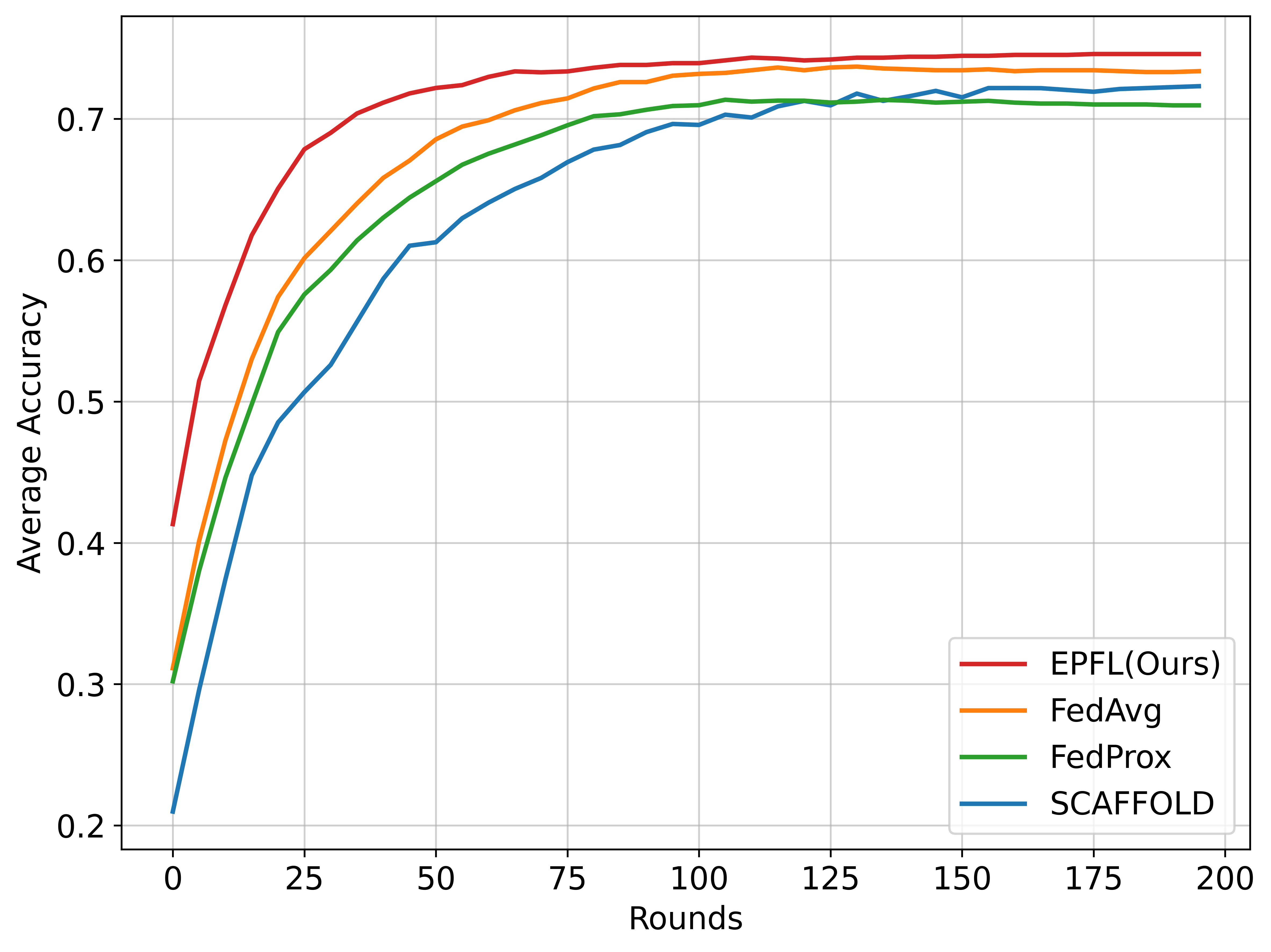}
    \caption{Convergence of Average Accuracy Over Training Rounds} 
    \label{avg_acc} 
\end{figure}

\subsection{parameter sensitivity}

To evaluate the impact of parameter selection on similarity computation, we varied the hyperparameter \( \psi \) to test different subsets of LoRA-B matrix parameters from the ViT model. Specifically, we selected the first half, the second half, and the entire set of parameters for similarity computation. The corresponding results are presented in Figure \ref{abaltion}(b).

The first half consistently performed the worst across all datasets, indicating insufficient information capture for effective similarity computation. Meanwhile, the second half and the entire set showed dataset-dependent performance, suggesting that the optimal parameter subset varies with dataset characteristics. This underscores the importance of selecting appropriate parameter subsets for specific tasks.

\begin{figure}[htb] 
    \centering 
    \includegraphics[width=0.35\textwidth]{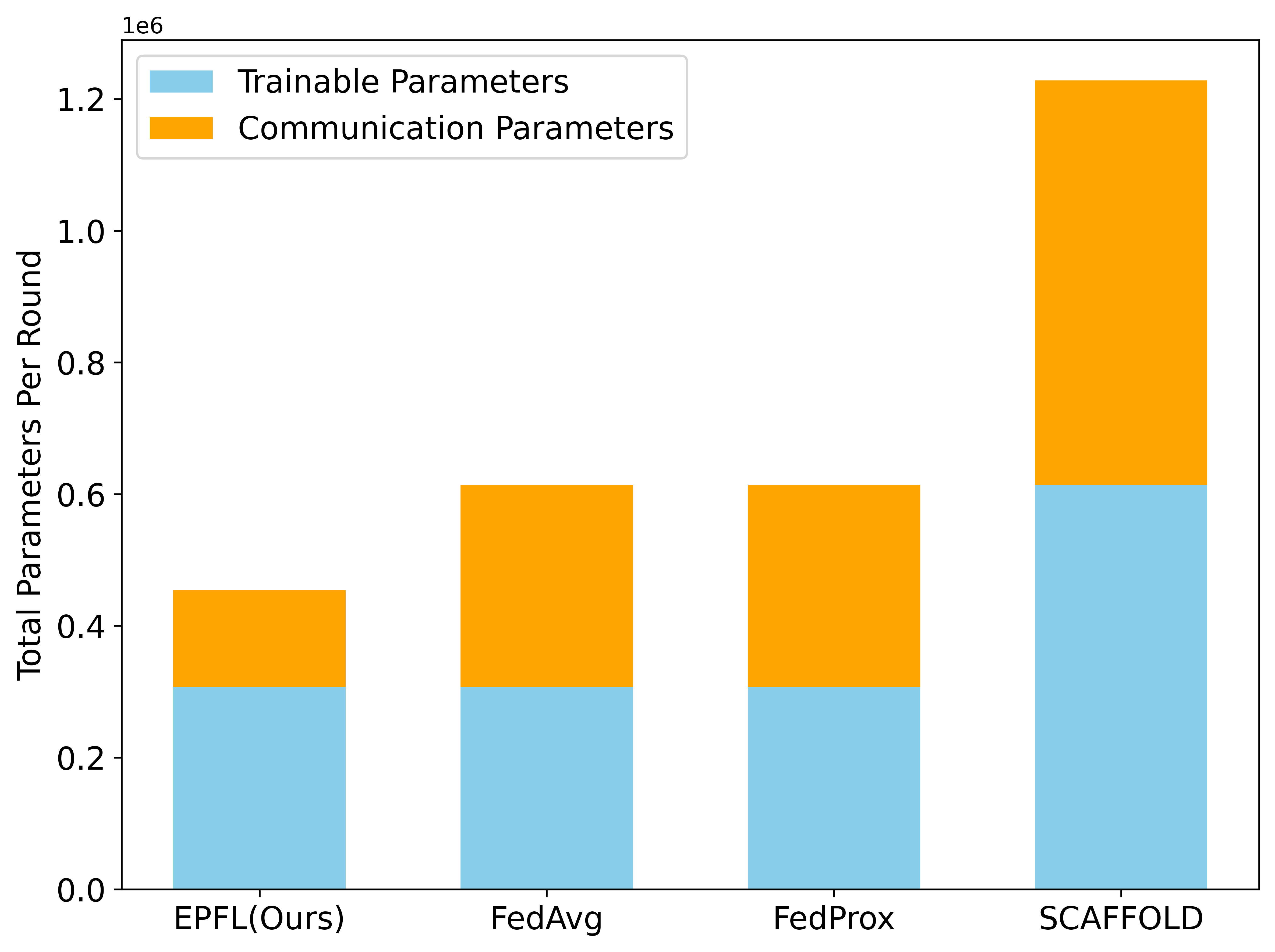}
    \caption{Parameter Efficiency Comparison} 
    \label{para} 
\end{figure}

\subsection{Efficiency Analysis}\label{analysis}
To further demonstrate the efficiency of our method, we evaluated it in terms of convergence speed and parameter efficiency:

\textbf{Convergence Speed: }We compared the average client accuracy over training rounds on the OrganCMNIST dataset. The results, shown in Figure \ref{avg_acc}, indicate that our method converges faster and achieves higher final accuracy compared to other baseline algorithms, such as FedAvg and FedProx. This demonstrates the computational efficiency of our similarity-weighted aggregation mechanism.

\textbf{Parameter Efficiency:} To quantify the parameter efficiency of our method, we compared the Total Parameters per Round (including trainable and communication parameters) across different algorithms. The results, presented in Figure \ref{para}, highlight that our method significantly reduces communication overhead while maintaining competitive performance. By combining faster convergence with reduced parameter usage, our method demonstrates clear efficiency advantages, making it particularly suitable for resource-constrained and privacy-sensitive FL scenarios.

\section{Conclusions}
We introduced FHBench, a comprehensive benchmark tailored to the unique challenges of healthcare FL, including multimodal diagnostics, non-IID distributions, and privacy constraints. Extensive evaluations of FHBench highlight its utility in reflecting real-world medical scenarios. Building on this foundation, we proposed EPFL, a PFL framework leveraging Low-Rank Adaptation (LoRA) for efficient and adaptive parameter aggregation. EPFL demonstrated superior performance and efficiency across diverse medical modalities, addressing the limitations of existing methods. While this study establishes a robust benchmark and method, future work will expand FHBench to include additional tasks and modalities, further advancing FL in complex and resource-constrained healthcare settings.

\newpage
\bibliographystyle{named}
\bibliography{ref}

\newpage
\appendix
\section{Dataset Details and Preprocessing}
\label{datasets}
\textbf{MedMnist}: This is a large-scale MNIST-like collection of standardized biomedical images. It includes 12 datasets for 2D images and 6 datasets for 3D images. We have selected high-resolution versions (224x224) of OrganAMNIST, OrganCMNIST, and OrganSMNIST.These datasets contain resampled slices from 3D CT images, captured in axial, coronal, and sagittal planes respectively, each encompassing 11 classes. The sample sizes are as follows: OrganAMNIST with 58,850 samples, OrganCMNIST with 23,660 samples, and OrganSMNIST with 25,221 samples.

\textbf{Medical Abstracts Text}: This dataset is a corpus of medical text classification, compiled from Kaggle. It consists of 14,438 medical abstracts, each categorized into one of five classes describing patient conditions. The distribution of classes includes 3,163 abstracts on neoplasms, 1,494 on digestive system diseases, 1,925 on nervous system diseases, 3,051 on cardiovascular diseases, and 4,805 on general pathological conditions.

\textbf{PubMed 200k}: Based on data from PubMed, this dataset is designed for sequential sentence classification, featuring 195,654 abstracts from randomized controlled trials. These abstracts total approximately 2.3 million sentences, distributed across five categories: Objective, Background, Conclusions, Methods, and Results. The dataset is randomly split into three subsets: a validation set containing 2,500 abstracts, a test set containing 2,500 abstracts, and a training set with the remaining 190,654 abstracts. A smaller version, PubMed 20k RCT, is also provided, comprising a training set of 15,000 abstracts, a validation set of 2,500 abstracts, and a test set of 2,500 abstracts. The 20k abstracts were selected from the larger dataset by prioritizing the most recently published studies, ensuring they represent contemporary research trends.

\textbf{ADFTD}: This public EEG time series dataset includes data from 36 Alzheimer's disease (AD) patients, 23 Frontotemporal Dementia (FTD) patients, and 29 healthy control (HC) subjects. It features 19 channels, and the raw sampling rate is 500Hz with 10uV/mm resolution. We followed the preprocessing steps described in \cite{wang2024medformer}, downsampling each trial to 256Hz and segmenting them into nonoverlapping 1-second samples, resulting in a total of 69,752 samples.

\textbf{PTBXL}: A large public ECG time series dataset recorded from 18,869 subjects, featuring 12 channels. The five labels represent four heart diseases and one healthy control category. The raw trials consist of 10-second time intervals, with sampling frequencies of 100Hz and 500Hz versions. Following the preprocessing steps described in \cite{wang2024medformer}, we utilize the 500Hz version downsample the data to 250Hz and normalize it using standard scalers, discarding any samples shorter than 1 second, resulting in 191,400 samples.

\textbf{ICBHI 2017}: Originating from the 2017 ICBHI challenge, this dataset includes 5.5 hours of recordings containing 6,898 respiratory cycles across 920 annotated audio samples from 126 subjects. The categories include 3,642 normal breathing cycles, 1,864 cycles with crackles, 886 with wheezes, and 506 with both crackles and wheezes. Following the preprocessing steps described in \cite{moummad2023resp}, we re-sampled all recordings to 16 kHz and limited the maximum duration of each respiratory cycle to 8 seconds.

\textbf{SPRSound}: An open-source pediatric respiratory sound database, this dataset includes 2,683 records and 9,089 respiratory sound events from 292 participants aged from 1 month to 18 years. It comprises 6,887 normal breathing cycles, 53 rhonchi, 865 wheezes, 17 stridor, 66 coarse crackles, 1,167 fine crackles, and 34 events featuring both crackle and wheeze. Consistent with the preprocessing approach in \cite{moummad2023resp}, all recordings were re-sampled to 16 kHz, and the maximum duration of each respiratory cycle was capped at 8 seconds.

\section{Methodologies and Implementation Details}\label{appendix method}

\textbf{FedAvg:}The foundational method for FL, where clients and the server collaboratively optimize a shared model by averaging local updates computed using the SGD optimizer. FedAvg is designed to be communication-efficient, allowing multiple local updates before aggregation. By avoiding the sharing of raw data, it ensures privacy while enabling collaborative model training.

\textbf{FedProx:}Addresses statistical heterogeneity by introducing an L2 regularization term that restricts local updates to remain close to the global model, improving stability during training. This proximal term restricts local updates from deviating too far from the global model, ensuring more stable training and improved convergence in non-IID settings.

\textbf{SCAFFOLD:}A FL algorithm that mitigates client drift caused by data heterogeneity by using control variates (gradients) at both the server and clients. These variates correct local updates, ensuring consistency across iterations and accelerating convergence. This approach is especially effective in non-IID settings, reducing communication costs and improving overall performance.

\textbf{APFL:}A FL algorithm that balances global and local model adaptation by learning a weighted combination of a shared global model and a personalized local model. This approach allows each client to adaptively adjust the mixture of global and local models based on its unique data distribution, improving performance on non-IID datasets while maintaining communication efficiency.

\textbf{APPLE:}A PFL method that allows clients to adaptively learn from other clients' models using a Directed Relationship (DR) vector while maintaining privacy by sharing only core models. APPLE dynamically balances global and local objectives, improving personalization and efficiency.

\textbf{FedALA:}introduces an Adaptive Local Aggregation (ALA) module that combines global and local models on each client to address data heterogeneity. It captures useful information from the global model element-wise, ensuring that the global model adapts better to client-specific data distributions.
\begin{table*}[t]
\centering
\begin{tabular}{cccccccc}
\hline
\textbf{Client} & \textbf{fedavg} & \textbf{fedprox} & \textbf{SCAFFOLD} & \textbf{APFL} & \textbf{APPLE} & \textbf{FedALA} & \textbf{EPFL(Ours)} \\ \hline
1               & 91.76           & 92.86            & 87.91             & 87.36         & 92.86          & 93.41           & 93.96               \\
2               & 87.63           & 85.48            & 82.80              & 79.57         & 86.02          & 88.17           & 88.71               \\
3               & 83.52           & 82.97            & 79.12             & 84.07         & 85.71          & 85.16           & 88.46               \\
4               & 91.76           & 91.76            & 92.31             & 88.46         & 91.76          & 93.96           & 92.31               \\
5               & 94.02           & 91.85            & 88.59             & 88.04         & 94.57          & 92.93           & 91.85               \\
6               & 85.41           & 81.62            & 80.54             & 76.76         & 88.11          & 84.86           & 83.78               \\
7               & 88.04           & 88.59            & 86.96             & 87.50          & 89.13          & 90.22           & 92.39               \\
8               & 82.61           & 83.15            & 82.07             & 73.37         & 84.24          & 83.15           & 77.72               \\
9               & 83.06           & 86.89            & 84.70              & 83.06         & 86.34          & 86.34           & 88.52               \\
10              & 92.39           & 90.22            & 86.96             & 88.59         & 91.85          & 91.85           & 93.48               \\
11              & 87.10            & 88.71            & 86.02             & 84.41         & 88.71          & 87.63           & 91.94               \\
12              & 85.79           & 87.43            & 85.25             & 80.87         & 87.43          & 88.52           & 88.52               \\
13              & 85.16           & 80.22            & 84.07             & 75.82         & 84.07          & 85.71           & 81.32               \\
14              & 96.65           & 97.77            & 94.97             & 96.65         & 98.32          & 97.77           & 98.32               \\
15              & 87.98           & 89.07            & 86.89             & 84.70          & 87.98          & 89.62           & 91.80                \\
16              & 87.57           & 92.97            & 87.57             & 85.41         & 87.57          & 88.11           & 89.73               \\
17              & 95.65           & 97.28            & 96.74             & 95.65         & 96.74          & 96.74           & 97.83               \\
18              & 82.16           & 82.70             & 84.32             & 75.68         & 81.62          & 82.70            & 77.84               \\
19              & 89.56           & 85.16            & 84.07             & 83.52         & 86.81          & 86.26           & 85.71               \\
20              & 90.32           & 89.78            & 87.10              & 85.48         & 90.86          & 91.40            & 88.71               \\
avg             & 86.50            & 86.70             & 85.00                & 83.14         & 87.60           & 86.94           & 88.13               \\ \hline
\end{tabular}
\caption{Accuracy for each client on OrganAMNIST.}

\end{table*}

\section{Detailed Results}
In this section, we present the detailed results for each client in a single experimental run. The validation accuracy achieved by each client, along with the overall mean accuracy, is summarized to provide a comprehensive view of the experiment's performance.
The experiments are conducted with 20 clients over 200 training rounds, using cross-entropy loss and SGD optimizer. The evaluation is performed on the FHBench dataset, which provides a comprehensive benchmark for assessing algorithms across multiple modalities. Each client is assigned a unique subset of the dataset, reflecting real-world non-IID data distributions. The experimental results are presented in the tables below, which compare the performance of various algorithms.

\begin{table*}[t]
\centering
\begin{tabular}{cccccccc}
\hline
\textbf{Client} & \textbf{fedavg} & \textbf{fedprox} & \textbf{SCAFFOLD} & \textbf{APFL} & \textbf{APPLE} & \textbf{FedALA} & \textbf{EPFL(Ours)} \\ \hline
1               & 77.63           & 75.00               & 76.32             & 73.68         & 75.00             & 73.68           & 75.00                  \\
2               & 70.51           & 73.08            & 67.95             & 76.92         & 70.51          & 74.36           & 73.08               \\
3               & 73.08           & 67.95            & 62.82             & 66.67         & 65.38          & 69.23           & 74.36               \\
4               & 82.67           & 84.00               & 70.67             & 78.67         & 86.67          & 78.67           & 85.33               \\
5               & 68.83           & 74.03            & 75.32             & 71.43         & 79.22          & 68.83           & 80.52               \\
6               & 80.77           & 79.49            & 83.33             & 76.92         & 79.49          & 85.90            & 87.18               \\
7               & 64.56           & 64.56            & 67.09             & 59.49         & 68.35          & 65.82           & 67.09               \\
8               & 82.05           & 80.77            & 78.21             & 76.92         & 85.90           & 83.33           & 82.05               \\
9               & 72.97           & 68.92            & 68.92             & 77.03         & 70.27          & 66.22           & 75.68               \\
10              & 67.90            & 80.25            & 75.31             & 59.26         & 69.14          & 62.96           & 72.84               \\
11              & 71.62           & 75.68            & 71.62             & 70.27         & 74.32          & 66.22           & 78.38               \\
12              & 77.50            & 75.00               & 78.75             & 70.00            & 73.75          & 80.00              & 75.00                  \\
13              & 77.63           & 77.63            & 81.58             & 68.42         & 78.95          & 84.21           & 76.32               \\
14              & 72.73           & 72.73            & 71.43             & 72.73         & 67.53          & 72.73           & 74.03               \\
15              & 64.47           & 65.79            & 63.16             & 69.74         & 60.53          & 68.42           & 71.05               \\
16              & 77.92           & 77.92            & 75.32             & 70.13         & 84.42          & 84.42           & 79.22               \\
17              & 89.74           & 83.33            & 80.77             & 75.64         & 79.49          & 82.05           & 82.05               \\
18              & 63.64           & 66.23            & 66.23             & 68.83         & 68.83          & 66.23           & 68.83               \\
19              & 72.37           & 64.47            & 60.53             & 64.47         & 68.42          & 64.47           & 67.11               \\
20              & 79.75           & 82.28            & 78.48             & 73.42         & 82.28          & 77.22           & 82.28               \\
avg             & 75.51           & 74.61            & 74.25             & 73.47         & 76.09          & 76.09           & 79.55               \\ \hline
\end{tabular}
\caption{Accuracy for each client on OrganCMNIST.}

\end{table*}

\begin{table*}[t]
\centering
\begin{tabular}{cccccccc}
\hline
\textbf{Client} & \textbf{fedavg} & \textbf{fedprox} & \textbf{SCAFFOLD} & \textbf{APFL} & \textbf{APPLE} & \textbf{FedALA} & \textbf{EPFL(Ours)} \\ \hline
1               & 64.63           & 68.29            & 64.63             & 67.07         & 64.63          & 67.07           & 67.07               \\
2               & 73.17           & 71.95            & 74.39             & 63.41         & 71.95          & 73.17           & 68.29               \\
3               & 76.19           & 71.43            & 73.81             & 66.67         & 79.76          & 77.38           & 71.43               \\
4               & 69.05           & 67.86            & 64.29             & 63.10          & 65.48          & 71.43           & 71.43               \\
5               & 53.09           & 59.26            & 59.26             & 61.73         & 58.02          & 59.26           & 66.67               \\
6               & 74.12           & 77.65            & 80.00               & 70.59         & 77.65          & 76.47           & 74.12               \\
7               & 74.70            & 72.29            & 71.08             & 72.29         & 73.49          & 73.49           & 78.31               \\
8               & 64.71           & 69.41            & 70.59             & 51.76         & 72.94          & 70.59           & 60.00                  \\
9               & 72.84           & 74.07            & 76.54             & 76.54         & 74.07          & 71.60            & 82.72               \\
10              & 69.14           & 69.14            & 72.84             & 66.67         & 72.84          & 70.37           & 72.84               \\
11              & 51.81           & 67.47            & 57.83             & 59.04         & 63.86          & 69.88           & 60.24               \\
12              & 77.50            & 76.25            & 70.00                & 66.25         & 76.25          & 76.25           & 72.5                \\
13              & 79.27           & 81.71            & 80.49             & 68.29         & 82.93          & 84.15           & 80.49               \\
14              & 62.35           & 60.00               & 63.53             & 60.00            & 65.88          & 54.12           & 64.71               \\
15              & 81.71           & 75.61            & 79.27             & 76.83         & 80.49          & 89.02           & 80.49               \\
16              & 62.35           & 61.18            & 58.82             & 56.47         & 67.06          & 62.35           & 64.71               \\
17              & 80.49           & 79.27            & 76.83             & 75.61         & 79.27          & 84.15           & 76.83               \\
18              & 72.62           & 72.62            & 75.00                & 59.52         & 73.81          & 69.05           & 69.05               \\
19              & 87.65           & 87.65            & 83.95             & 87.65         & 90.12          & 86.42           & 88.89               \\
20              & 53.49           & 58.14            & 51.16             & 56.98         & 52.33          & 54.65           & 59.30                \\
avg             & 67.39           & 66.83            & 66.97             & 67.00            & 69.27          & 70.39           & 70.69               \\ \hline
\end{tabular}
\caption{Accuracy for each client on OrganSMNIST.}
\end{table*}

\begin{table*}[t]
\centering
\begin{tabular}{cccccccc}
\hline
\textbf{Client} & \textbf{fedavg} & \textbf{fedprox} & \textbf{SCAFFOLD} & \textbf{APFL} & \textbf{APPLE} & \textbf{FedALA} & \textbf{EPFL(Ours)} \\ \hline
1               & 63.96           & 62.16            & 63.06             & 41.44         & 64.86          & 70.27           & 65.77               \\
2               & 67.26           & 67.26            & 67.26             & 46.90          & 63.72          & 66.37           & 65.49               \\
3               & 81.98           & 83.78            & 79.28             & 50.45         & 81.08          & 82.88           & 83.78               \\
4               & 77.27           & 75.45            & 71.82             & 68.18         & 70.91          & 77.27           & 80.91               \\
5               & 83.93           & 85.71            & 82.14             & 47.32         & 85.71          & 87.50            & 85.71               \\
6               & 60.36           & 57.66            & 49.55             & 43.24         & 53.15          & 63.06           & 57.66               \\
7               & 72.07           & 72.07            & 72.07             & 72.07         & 72.07          & 72.07           & 75.68               \\
8               & 84.68           & 84.68            & 82.88             & 59.46         & 83.78          & 84.68           & 81.98               \\
9               & 83.64           & 83.64            & 81.82             & 45.45         & 81.82          & 83.64           & 85.45               \\
10              & 63.39           & 61.61            & 63.39             & 58.93         & 61.61          & 67.86           & 63.39               \\
11              & 83.04           & 84.82            & 79.46             & 42.86         & 83.04          & 83.93           & 85.71               \\
12              & 63.72           & 64.60             & 63.72             & 57.52         & 62.83          & 66.37           & 59.29               \\
13              & 82.3            & 81.42            & 72.57             & 38.94         & 72.57          & 83.19           & 81.42               \\
14              & 83.04           & 79.46            & 83.04             & 68.75         & 79.46          & 81.25           & 75.89               \\
15              & 75.89           & 76.79            & 72.32             & 41.07         & 74.11          & 77.68           & 76.79               \\
16              & 69.03           & 67.26            & 66.37             & 66.37         & 69.03          & 69.03           & 66.37               \\
17              & 69.37           & 69.37            & 70.27             & 69.37         & 69.37          & 69.37           & 69.37               \\
18              & 81.82           & 79.09            & 75.45             & 60.00            & 78.18          & 80.91           & 87.27               \\
19              & 87.50            & 86.61            & 83.93             & 46.43         & 87.50           & 88.39           & 88.39               \\
20              & 60.18           & 60.18            & 60.18             & 59.29         & 60.18          & 61.06           & 59.29               \\
avg             & 72.98           & 73.13            & 72.05             & 55.06         & 71.48          & 73.69           & 74.76               \\ \hline
\end{tabular}
\caption{Accuracy for each client on Medical Abstracts Text.}
\end{table*}

\begin{table*}[t]
\centering
\begin{tabular}{cccccccc}
\hline
\textbf{Client} & \textbf{fedavg} & \textbf{fedprox} & \textbf{SCAFFOLD} & \textbf{APFL} & \textbf{APPLE} & \textbf{FedALA} & \textbf{EPFL(Ours)} \\ \hline
1               & 87.13           & 90.06            & 87.13             & 67.84         & 87.72          & 87.72           & 90.64               \\
2               & 90.06           & 88.89            & 88.89             & 59.65         & 88.30           & 88.89           & 91.81               \\
3               & 87.21           & 86.63            & 86.05             & 68.02         & 87.79          & 89.53           & 84.30                \\
4               & 96.45           & 95.86            & 94.67             & 66.86         & 95.86          & 96.45           & 96.45               \\
5               & 82.35           & 81.18            & 78.82             & 49.41         & 80.00             & 85.29           & 87.65               \\
6               & 95.32           & 94.74            & 94.15             & 66.67         & 94.15          & 95.32           & 93.57               \\
7               & 79.53           & 78.36            & 76.02             & 44.44         & 74.85          & 78.95           & 80.12               \\
8               & 92.44           & 92.44            & 92.44             & 63.95         & 93.02          & 92.44           & 95.35               \\
9               & 88.30            & 88.30             & 90.06             & 46.78         & 88.89          & 90.06           & 90.06               \\
10              & 78.49           & 77.33            & 76.74             & 48.26         & 77.91          & 78.49           & 78.49               \\
11              & 77.78           & 76.61            & 76.61             & 49.12         & 76.61          & 77.78           & 75.44               \\
12              & 93.60            & 94.19            & 94.77             & 66.28         & 92.44          & 94.77           & 93.60                \\
13              & 80.23           & 77.91            & 75.58             & 45.35         & 80.23          & 81.40            & 82.56               \\
14              & 93.02           & 91.86            & 93.60              & 58.72         & 91.86          & 93.60            & 90.12               \\
15              & 91.23           & 90.06            & 92.98             & 59.65         & 90.64          & 90.06           & 92.4                \\
16              & 76.16           & 77.33            & 71.51             & 51.74         & 76.16          & 76.74           & 74.42               \\
17              & 92.98           & 90.06            & 92.98             & 66.67         & 90.06          & 92.40            & 89.47               \\
18              & 77.91           & 76.16            & 72.67             & 47.67         & 74.42          & 79.07           & 75.00                 \\
19              & 80.23           & 78.49            & 79.07             & 48.84         & 79.07          & 81.98           & 80.81               \\
20              & 94.19           & 93.60             & 94.19             & 43.02         & 93.02          & 94.77           & 91.28               \\
avg             & 85.58           & 84.38            & 83.29             & 56.54         & 84.26          & 85.70            & 86.18               \\ \hline
\end{tabular}
\caption{Accuracy for each client on PubMed200k RCT.}
\end{table*}

\begin{table*}[t]
\centering
\begin{tabular}{cccccccc}
\hline
\textbf{Client} & \textbf{fedavg} & \textbf{fedprox} & \textbf{SCAFFOLD} & \textbf{APFL} & \textbf{APPLE} & \textbf{FedALA} & \textbf{EPFL(Ours)} \\ \hline
1               & 48.57           & 48.57            & 48.57             & 48.57         & 48.57          & 45.71           & 51.43               \\
2               & 100.0             & 100.0              & 100.0              & 100.0           & 100.0            & 100.0             & 100.0                 \\
3               & 100.0             & 100.0              & 100.0              & 100.0           & 100.0            & 100.0             & 100.0                 \\
4               & 51.43           & 51.43            & 51.43             & 51.43         & 48.57          & 42.86           & 51.43               \\
5               & 100.0             & 100.0              & 100.0              & 100.0           & 100.0            & 100.0             & 100.0       \\
6               & 17.65           & 17.65            & 23.53             & 44.12         & 44.12          & 38.24           & 47.06               \\
7               & 51.43           & 45.71            & 51.43             & 51.43         & 54.29          & 51.43           & 51.43               \\
8               & 100.0             & 100.0              & 100.0              & 100.0           & 100.0            & 100.0             & 100.0                 \\
9               & 100.0             & 100.0              & 100.0              & 100.0           & 100.0            & 100.0             & 100.0                 \\
10              & 58.82           & 58.82            & 58.82             & 58.82         & 55.88          & 55.88           & 58.82               \\
11              & 26.47           & 38.24            & 44.12             & 44.12         & 44.12          & 41.18           & 44.12               \\
12              & 100.0             & 100.0              & 100.0              & 100.0           & 100.0            & 100.0             & 100.0                 \\
13              & 54.55           & 51.52            & 54.55             & 54.55         & 51.52          & 48.48           & 54.55               \\
14              & 100.0             & 100.0              & 100.0              & 100.0           & 100.0            & 100.0             & 100.0                 \\
15              & 30.30            & 24.24            & 27.27             & 63.64         & 63.64          & 60.61           & 63.64               \\
16              & 69.70            & 69.70             & 69.70              & 69.70          & 69.70           & 69.70            & 69.70                \\
17              & 17.14           & 42.86            & 20.00                & 51.43         & 45.71          & 42.86           & 45.71               \\
18              & 93.94           & 93.94            & 93.94             & 93.94         & 93.94          & 93.94           & 93.94               \\
19              & 52.94           & 52.94            & 52.94             & 61.76         & 50.00             & 41.18           & 55.88               \\
20              & 96.97           & 96.97            & 96.97             & 96.97         & 96.97          & 96.97           & 96.97               \\
avg             & 68.48           & 72.82            & 71.4              & 75.47         & 74.11          & 73.42           & 75.99               \\ \hline
\end{tabular}
\caption{Accuracy for each client on ICBHI.}
\end{table*}

\begin{table*}[t]
\centering
\begin{tabular}{cccccccc}
\hline
\textbf{Client} & \textbf{fedavg} & \textbf{fedprox} & \textbf{SCAFFOLD} & \textbf{APFL} & \textbf{APPLE} & \textbf{FedALA} & \textbf{EPFL(Ours)} \\ \hline
1               & 51.06           & 53.19            & 53.19             & 51.06         & 51.06          & 51.06           & 51.06               \\
2      & 100.0             & 100.0              & 100.0              & 100.0           & 100.0            & 100.0             & 100.0                 \\
3               & 78.57           & 78.57            & 78.57             & 78.57         & 78.57          & 78.57           & 78.57               \\
4               & 71.11           & 75.56            & 71.11             & 71.11         & 71.11          & 71.11           & 71.11               \\
5              & 100.0             & 100.0              & 100.0              & 100.0           & 100.0            & 100.0             & 100.0                 \\
6               & 50.00              & 50.00               & 50.00                & 50.00            & 50.00             & 50.00              & 50.00                  \\
7                & 100.0             & 100.0              & 100.0              & 100.0           & 100.0            & 100.0             & 100.0                 \\
8               & 50.00              & 50.00               & 50.00                & 36.96         & 52.17          & 52.17           & 52.17               \\
9               & 51.11           & 51.11            & 51.11             & 42.22         & 51.11          & 51.11           & 51.11               \\
10              & 100.0             & 100.0              & 100.0              & 100.0           & 100.0            & 100.0             & 100.0                 \\
11              & 51.11           & 51.11            & 51.11             & 46.67         & 51.11          & 51.11           & 51.11               \\
12               & 100.0             & 100.0              & 100.0              & 100.0           & 100.0            & 100.0             & 100.0                 \\
13              & 55.56           & 55.56            & 55.56             & 57.78         & 55.56          & 57.78           & 55.56               \\
14               & 100.0             & 100.0              & 100.0              & 100.0           & 100.0            & 100.0             & 100.0                 \\
15              & 50.00              & 50.00               & 50.00                & 50.00            & 50.00             & 50.00              & 50.00                  \\
16              & 100.0             & 100.0              & 100.0              & 100.0           & 100.0            & 100.0             & 100.0                 \\
17              & 48.94           & 48.94            & 51.06             & 19.15         & 48.94          & 48.94           & 48.94               \\
18             & 100.0             & 100.0              & 100.0              & 100.0           & 100.0            & 100.0             & 100.0                 \\
19              & 54.35           & 58.70             & 56.52             & 54.35         & 54.35          & 56.52           & 56.52               \\
20               & 100.0             & 100.0              & 100.0              & 100.0           & 100.0            & 100.0             & 100.0                 \\
avg             & 77.17           & 77.29            & 77.43             & 73.93         & 77.32          & 77.39           & 78.45               \\ \hline
\end{tabular}
\centering
\caption{Accuracy for each client on SPRS.}
\end{table*}

\begin{table*}[t]
\centering
\begin{tabular}{cccccccc}
\hline
\textbf{Client} & \textbf{fedavg} & \textbf{fedprox} & \textbf{SCAFFOLD} & \textbf{APFL} & \textbf{APPLE} & \textbf{FedALA} & \textbf{EPFL(Ours)} \\ \hline
1               & 69.95           & 68.54            & 69.95             & 69.95         & 69.95          & 69.95           & 69.95               \\
2               & 50.00              & 50.00               & 50.00               & 53.30          & 41.51          & 50.00              & 55.66               \\
3               & 84.43           & 84.43            & 84.43             & 84.43         & 77.36          & 84.43           & 84.43               \\
4               & 61.14           & 60.66            & 61.14             & 61.14         & 61.14          & 61.14           & 61.14               \\
5               & 62.86           & 62.38            & 62.86             & 62.86         & 62.86          & 62.86           & 62.86               \\
6               & 84.36           & 84.36            & 84.36             & 84.36         & 74.88          & 84.36           & 84.36               \\
7               & 78.20            & 77.25            & 77.25             & 77.25         & 71.09          & 77.25           & 77.25               \\
8               & 50.71           & 48.82            & 50.71             & 54.98         & 50.71          & 51.18           & 51.18               \\
9               & 48.11           & 47.64            & 48.11             & 53.30          & 48.11          & 49.06           & 52.83               \\
10              & 78.20            & 77.73            & 78.20              & 78.20          & 72.04          & 78.20            & 78.20                \\
11              & 61.61           & 61.14            & 61.61             & 61.61         & 61.61          & 61.61           & 61.61               \\
12              & 81.13           & 80.66            & 79.25             & 81.13         & 67.92          & 81.13           & 81.13               \\
13              & 82.16           & 82.16            & 72.77             & 82.63         & 71.36          & 82.63           & 82.63               \\
14              & 55.71           & 54.76            & 55.71             & 55.71         & 55.71          & 55.71           & 56.19               \\
15              & 67.77           & 67.30             & 67.77             & 67.77         & 67.77          & 67.77           & 67.77               \\
16              & 81.60            & 80.66            & 81.60              & 81.60          & 71.23          & 81.60            & 81.60                \\
17              & 43.6            & 42.18            & 40.28             & 34.12         & 40.28          & 40.76           & 40.76               \\
18              & 47.17           & 45.28            & 43.87             & 44.81         & 42.45          & 44.81           & 44.81               \\
19              & 68.87           & 66.98            & 68.87             & 68.87         & 68.87          & 68.87           & 68.87               \\
20              & 78.10            & 78.10             & 66.19             & 78.57         & 70.48          & 78.57           & 78.57               \\
avg             & 66.66           & 66.18            & 65.72             & 66.64         & 62.69          & 66.78           & 67.05               \\ \hline
\end{tabular}
\caption{Accuracy for each client on ADFTD.}
\end{table*}

\begin{table*}[t]
\centering
\begin{tabular}{cccccccc}
\hline
\textbf{Client} & \textbf{fedavg} & \textbf{fedprox} & \textbf{SCAFFOLD} & \textbf{APFL} & \textbf{APPLE} & \textbf{FedALA} & \textbf{EPFL(Ours)} \\ \hline
1               & 67.01           & 67.70             & 63.92             & 66.67         & 66.67          & 67.35           & 68.04               \\
2               & 87.15           & 88.19            & 88.54             & 86.81         & 88.19          & 88.89           & 93.06               \\
3               & 59.25           & 62.33            & 59.93             & 59.93         & 62.33          & 61.99           & 63.36               \\
4               & 89.62           & 89.27            & 90.66             & 87.20          & 90.66          & 90.66           & 90.66               \\
5               & 92.39           & 92.73            & 93.08             & 89.97         & 93.43          & 93.77           & 94.12               \\
6               & 55.48           & 60.62            & 57.88             & 57.19         & 58.22          & 56.85           & 55.48               \\
7               & 65.17           & 71.03            & 66.21             & 65.52         & 70.69          & 65.86           & 68.28               \\
8               & 90.72           & 90.03            & 92.78             & 90.03         & 90.72          & 90.38           & 92.10                \\
9               & 95.16           & 94.46            & 93.77             & 91.00            & 96.19          & 95.50            & 94.46               \\
10              & 57.39           & 61.51            & 51.55             & 53.61         & 57.73          & 56.70            & 59.79               \\
11              & 59.25           & 58.56            & 54.11             & 61.64         & 61.99          & 57.88           & 60.27               \\
12              & 88.62           & 87.59            & 86.55             & 86.21         & 87.59          & 86.90            & 89.66               \\
13              & 62.46           & 65.53            & 62.80              & 62.12         & 62.46          & 63.14           & 65.87               \\
14              & 82.07           & 81.03            & 78.62             & 72.76         & 82.41          & 83.10            & 82.07               \\
15              & 98.27           & 97.58            & 97.58             & 97.23         & 97.92          & 98.96           & 97.23               \\
16              & 56.90            & 61.38            & 52.07             & 54.14         & 58.97          & 56.55           & 56.55               \\
17              & 64.48           & 61.72            & 62.41             & 63.10          & 63.79          & 62.07           & 62.76               \\
18              & 85.81           & 83.39            & 84.78             & 83.04         & 84.78          & 86.16           & 87.89               \\
19              & 65.64           & 67.01            & 65.29             & 65.64         & 67.01          & 68.38           & 67.01               \\
20              & 85.42           & 85.42            & 85.76             & 83.68         & 88.19          & 87.15           & 86.81               \\
avg             & 75.04           & 75.37            & 73.91             & 73.34         & 75.19          & 75.67           & 75.91               \\ \hline
\end{tabular}
\caption{Accuracy for each client on PTB-XL.}
\end{table*}

\section{Discussion}
Our proposed method demonstrates superior overall performance across most datasets, highlighting its effectiveness in addressing the challenges posed by non-IID data in real healthcare scenarios. These results validate the robustness of our approach in mitigating data heterogeneity issues.
In pursuit of computational efficiency and to better simulate real-world scenarios with limited data availability, the data allocated to each client was kept minimal. While this approach ensured faster processing and aligned with the characteristics of practical federated learning settings, it also led to some performance discrepancies and fairness issues among clients. This underscores the importance of exploring more fine-grained data partitioning and preprocessing strategies to strike a better balance between efficiency, realism, and equitable performance distribution. We plan to address these aspects in our future work to further enhance the robustness and fairness of our method.

\end{document}